\documentclass{article}
\usepackage{microtype}
\usepackage{graphicx}
\usepackage{subfigure}
\usepackage{booktabs} 

\usepackage{hyperref}

\usepackage[accepted]{icml2023}

\usepackage{amsmath}
\usepackage{amssymb}
\usepackage{mathtools}
\usepackage{amsthm}
\usepackage{enumerate}

\usepackage{array} 

\usepackage{longtable}

\theoremstyle{plain}

\theoremstyle{definition}

\theoremstyle{remark}

\newcommand{\alg}{OPA}
\newcommand{\refEq}[1]{Eq.(\ref{#1})}
\newcommand{\reffig}[1]{Figure \ref{#1}}
\newcommand{\reftab}[1]{Table \ref{#1}}
\newcommand{\refsec}[1]{Section \ref{#1}}
\newcommand{\refapp}[1]{Appendix \ref{#1}}

\newcommand{\ms}[2]{#1 {\tiny $\pm$ #2}}
\newcommand{\ustc}{University of Science and Technology of China}
\newcommand{\ucas}{University of Chinese Academy of Sciences, China}
\newcommand{\skl}{SKL of Processors, Institute of Computing Technology, CAS}
\newcommand{\camb}{Cambricon Technologies}

\newcommand{\soft}{Institute of Software Chinese Academy of Sciences}

\newcommand{\dsrc}{$\mathcal{M}_S$}
\newcommand{\dtarg}{$\mathcal{M}_T$}

\newcommand{\itembg}[1]{\includegraphics[width=0.02\textwidth]{figures/env_img_ver#1/item_0.png}}
\newcommand{\itemzombie}[1]{\includegraphics[width=0.02\textwidth]{figures/env_img_ver#1/item_1.png}}
\newcommand{\itemagent}[1]{\includegraphics[width=0.02\textwidth]{figures/env_img_ver#1/item_2.png}}
\newcommand{\itemcow}[1]{\includegraphics[width=0.02\textwidth]{figures/env_img_ver#1/item_3.png}}
\newcommand{\itemwall}[1]{\includegraphics[width=0.02\textwidth]{figures/env_img_ver#1/item_4.png}}

\usepackage[textsize=tiny]{todonotes}

\icmltitlerunning{Online Prototype Alignment for Few-shot Policy Transfer}

\begin{document}

\twocolumn[
\icmltitle{Online Prototype Alignment for Few-shot Policy Transfer}




\icmlsetsymbol{equal}{*}

\begin{icmlauthorlist}
\icmlauthor{Qi~Yi}{ustc,skl,camb}
\icmlauthor{Rui~Zhang}{skl}
\icmlauthor{Shaohui~Peng}{skl,ucas,camb}
\icmlauthor{Jiaming~Guo}{skl,ucas,camb}
\icmlauthor{Yunkai~Gao}{ustc,skl,camb}
\icmlauthor{Kaizhao~Yuan}{skl,ucas,camb}
\icmlauthor{Ruizhi~Chen}{soft}
\icmlauthor{Siming~Lan}{ustc,skl,camb}
\icmlauthor{Xing~Hu}{skl}
\icmlauthor{Zidong~Du}{skl}
\icmlauthor{Xishan~Zhang}{skl,camb}
\icmlauthor{Qi~Guo}{skl}
\icmlauthor{Yunji~Chen}{skl,ucas}
\end{icmlauthorlist}

\icmlaffiliation{ustc}{\ustc{}}
\icmlaffiliation{skl}{\skl{}}
\icmlaffiliation{ucas}{\ucas{}}
\icmlaffiliation{camb}{\camb{}}
\icmlaffiliation{soft}{\soft{}}
\icmlcorrespondingauthor{Yunji~Chen}{cyj@ict.ac.cn}

\icmlkeywords{Machine Learning, ICML}

\vskip 0.3in
]



\printAffiliationsAndNotice{}  

\begin{abstract}
Domain adaptation in reinforcement learning (RL) mainly deals with the changes of observation when transferring the policy to a new environment.
Many traditional approaches of domain adaptation in RL manage to learn a mapping function between the source and target domain in explicit or implicit ways. 
However, they typically require access to abundant data from the target domain. Besides, they often rely on visual clues to learn the mapping function and may fail when the source domain looks quite different from the target domain.
To address these problems, we propose a novel framework Online Prototype Alignment (OPA)  to learn the mapping function based on the functional similarity of elements and is able to achieve the few-shot policy transfer within only several episodes. 
The key insight of OPA is to introduce an exploration mechanism that can interact with the unseen elements of the target domain in an efficient and purposeful manner, and then connect them with the seen elements in the source domain according to their functionalities (instead of visual clues). 
Experimental results show that when the target domain looks visually different from the source domain, OPA can achieve better transfer performance even with much fewer samples from the target domain, outperforming prior methods. 
\end{abstract}

\section{Introduction} 
Deep Reinforcement Learning has achieved impressive results in many domains, such as Atari \cite{rl_1} and Mujoco \cite{rl_2}.
However, traditional RL algorithms typically require many interactions with the environment \cite{rl_data}. 
Besides, the learned policy can easily be over-fitted to the source domain where it is trained and may collapse if faced with slight changes in the target domain \cite{tsf_hard,concept_rl}. 
Therefore, it is essential to investigate how a policy can be transferred to a new environment. 

\begin{figure}[t]
\centering
\subfigure[Case A: Source]{
   \includegraphics[width=0.15\textwidth]{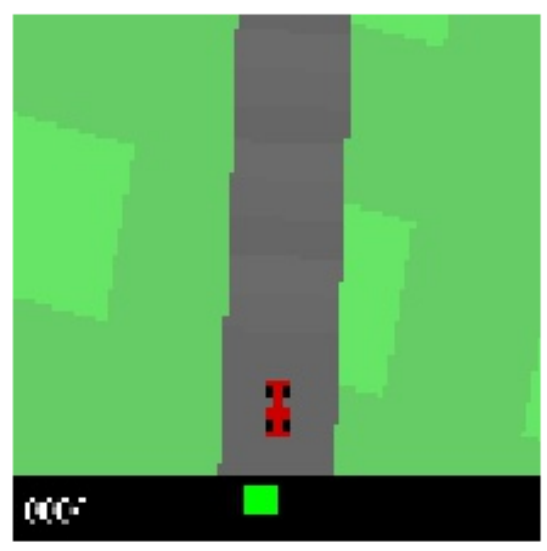}
}
\hspace{20pt}
\subfigure[Case A: Target]{
   \includegraphics[width=0.15\textwidth]{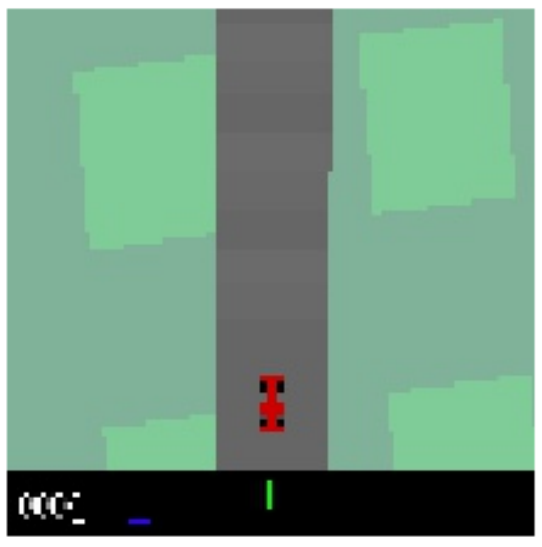}
}
\subfigure[Case B: Source]{
   \includegraphics[width=0.15\textwidth]{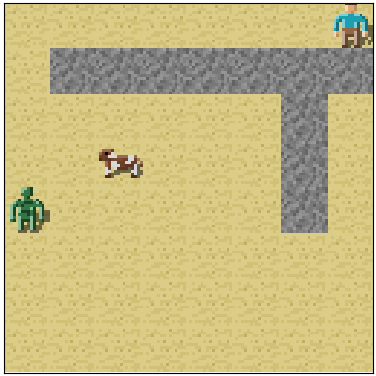}
}
\hspace{20pt}
\subfigure[ Case B: Target]{
   \includegraphics[width=0.15\textwidth]{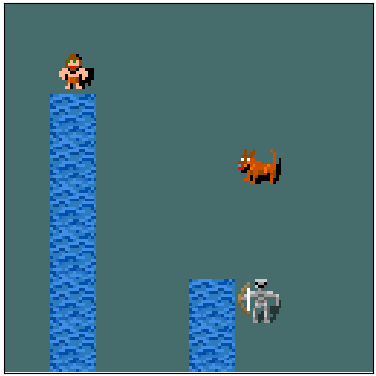}
}
\caption{
The source and target domain for cases A \cite{lusr} and B (considered in this work). 
Case B is more difficult than case A because we can not solely rely on visual clues to learn the mapping function between the source and target domain.
}
 \label{fig.obs_img}
\end{figure}

When trying to achieve such transfer, one of the most critical problems is dealing with the changes to the observation distribution, also known as domain adaptation in RL \cite{darla,dm_adapt}.
Many previous works try to solve this problem by learning a mapping function between the target and source domain. 
For example, \cite{tsf_img2img_1,tsf_img2img_2,tsf_img2img_3} learn an image-to-image translation model that can map the observations from the target domain back into the source domain, and therefore the policy trained in the source domain is directly applicable when equipped with such translation. 
Some works \cite{lusr,darla,tsf_dynamic} also learn such a mapping indirectly, in which the observations from the source and target domain are mapped into aligned representations.

Although these works have achieved compelling performance in many tasks, they typically require access to abundant data from the target domain, which can be problematic when collecting these data is expensive. 
Besides, most works are applicable when the target domain \emph{looks similar in appearances} to the source domain (e.g. case A in \reffig{fig.obs_img}).
When faced with more challenging cases where the elements in the target domain have the same underlying functionalities but irrelevant appearances (e.g. case B in \reffig{fig.obs_img}), these methods are likely to fail.
Therefore, how to quickly transfer between domains of irrelevant appearances still remains a problem.

On the other hand, it is possible for our human beings to achieve such a transfer.
This is because we can utilize the \emph{functional similarity} between elements to determine the mapping function between the source and target domain. 
For example, suppose we can get a score by eating an `apple' and then learn to seek and eat the apples in a game.
When faced with an unseen `pear' in a new game, we find that we can also get a score by eating the `pear', then we can quickly treat it as an `apple' and also seek and eat the pears in the new game. 
In summary, the policy transfer from an `apple' to a `pear' is based on the fact that `pear' and `apple' have the same underlying functionality, i.e. both eating a `pear' and an `apple' can increase the score.
However, learning the functional similarity of elements between source and target domains is difficult, because we have to interact \emph{actively} with those unseen elements in the target domain.
Thus, an efficient exploration mechanism is needed to discover the underlying functionalities.

Following the insight above, in this work, we propose a novel framework named \textbf{O}nline \textbf{P}rototype \textbf{A}lignment (\alg{}) to learn the mapping function based on the functional similarity of elements and achieve the few-shot policy transfer within only several episodes.
To represent the underlying functionalities of elements, we assume the elements in the tasks can be divided into several kinds of prototypes such that elements of the same prototype share the same functionalities.
To discover the prototypes of unseen elements quickly, 
\alg{} introduces an exploration policy. 
The exploration policy is trained by maximizing the mutual information between the trajectories it produces and the prototypes of unseen elements, therefore it can interact with these unseen elements in an efficient and purposeful manner to infer their prototypes.
When deployed on the target domain, \alg{} first distinguishes unseen elements by novelty detection. 
Then the exploration policy interacts with these unseen elements so that \alg{} can infer their prototypes based on the produced trajectories. 
Finally, by building a mapping function based on the discovered prototypes,
we can directly transfer the policy trained in the source domain to solve the task in the target domain.
Compared with previous works, \alg{} introduces an exploration mechanism to learn the mapping function based on the functional similarity between elements in the source and target domain, and can efficiently achieve few-shot policy transfer even if there are no visual clues for transfer between the two domains. 

The experiments are carried out on the task suite named Hunter \cite{ocarl}. 
To reveal the strength of \alg{}, we use the original version of Hunter as the source domain and derive a new variant that looks significantly different from the original as the target domain. 
Compared with several baselines, \alg{} can achieve better transfer performance by only using a few data from the target domain, outperforming other baselines. 

\section{Related Work}\label{sec.related}
\paragraph{Domain Adaptation in RL:} 
The goal of domain adaptation is to address the domain shift between the source and target domain. 
Most domain adaptation approaches are designed to deal with the changes to the observation distribution. 
Current domain adaptation methods can be roughly divided into three categories: domain randomization \cite{domain_random_1, domain_random_2,domain_random_3}, image-to-image translation \cite{tsf_img2img_1,tsf_img2img_2,tsf_img2img_3,tsf_img2img_4}, and adaptation via aligned representations \cite{lusr, darla, tsf_dynamic}. 

In domain randomization, a meta-simulator is required to generate many variants of the source domain. As a result, policies trained in these variants can learn to attend to the common features. However, these methods cannot work when the meta-simulator is not available, which is generally costly to attain in practice.
In image-to-image translation approaches, a mapping function is learned to map the pixel observations from the target domain to the source domain. 
Such mapping is often learned via generative adversarial networks (GANs). 
In adaptation approaches via aligned representations, the source and target domain observations are mapped into a well-regularized latent space. Ideally, representations in this latent space can share consistent semantic meanings no matter which domain they come from.
For example, \cite{lusr} explicitly splits the latent representations into domain-specific and domain-general features and then builds policy on the domain-general features to ignore domain-specific variations.

Although these works have achieved compelling performance, they typically require access to abundant data
from the target domain (or other domains that are different from the source domain). Besides, most of them rely on visual clues to learn the mapping function, which can be problematic when the elements in the target domain have irrelevant appearances.


\paragraph{Object Oriented RL:} The basic assumption of Object Oriented RL (OORL) is that the state space of MDPs can be represented in terms of objects, which is inspired by the fact that objects are the basic units of recognizing the world. In OORL, the agent's observations are a set of object representations, and the agent can solve the task by reasoning over these objects. By leveraging the invariance of objects' functionalities in different scenarios, policy trained in this way can often achieve better generalization ability \cite{ocarl,rrl}. Recent progress \cite{space, scalor} in Unsupervised Object Discovery also boosts the development of OORL. In our work, we follow the basic settings of OORL.

\section{Preliminaries} \label{sec.prel}
\subsection{Notation}
We assume the underlying environment is a Markov decision process (MDP), described by the tuple $\mathcal{M} = (S, A, P_T, R)$, where $S$ is the state space, $A$ the action space, $P_T: S_t\times A_t\times S_{t+1}\rightarrow [0,1]$ the transition probability function which determines the distribution of next state given current state and action, and $R: S_t\times A_t\times S_{t+1}\rightarrow \mathbb{R}$ the reward function. Given the current state $s\in S$, an agent chooses its action $a\in A$ according to a policy function $a\sim \pi(\cdot|s)$. This action will update the system state to a new state $s'$ according to the transition function $P_T$, and then a reward $r =R(s,a,s')\in \mathbb{R}$ is given to the agent. The goal of the agent is to maximize the expected cumulative rewards by learning a policy $\pi$:
\begin{equation}
    \begin{aligned}
        J(\pi) = \mathbb{E}_{\tau \sim \pi} \sum_{t=0}^T R(s_t, a_t, s_{t+1}),
    \end{aligned}
    \label{eq.rl_obj}
\end{equation}
where $\tau := (s_0,a_0,r_0,...,s_T)$ is the trajectory generated by $\pi$.

In this work, we also assume the state space $S$ can be broken into a set of object representations: $S=\prod_{i=1}^N O$, where $O$ is the space of object representations.

\subsection{Problem Statement}
We consider the domain adaptation problem in which a task policy $\pi_{task}$ is first trained in the source domain $\mathcal{M}_S=(S^{source}, A, P_T^{source}, R^{source})$ and then transferred to the target domain $\mathcal{M}_T = (S^{target}, A, P_T^{target}, R^{target})$. 
We also assume that the $\mathcal{M}_S$ and $\mathcal{M}_T$ share the \emph{same underlying dynamics and reward structures} such that there exists a mapping function $f: S^{target} \rightarrow S^{source}$ and $\pi_{task}$ can achieve optimal transfer performance when equipped with $f$ (i.e. $\pi_{task}\circ f$).

\begin{algorithm}[t]
   \caption{The training procedure of \alg{}}
   \label{alg.opa}
\begin{algorithmic}
   \STATE {\bfseries Input:} $\mathcal{M}_S$
    \STATE {\bfseries Output:} $\pi_{task}$, $\pi_{exp}$, $q_\theta$, $\Psi_{\texttt{IsUnseen}}$
    \STATE \textit{/* Train $\pi_{task}$ */}
    \STATE Train $\pi_{task}$ to solve $\mathcal{M}_S$, and save the historic trajectories as $D_{his}$.
    \STATE \textit{/* Train} $\Psi_{\texttt{IsUnseen}}$ \textit{*/}
    \STATE Train $g_{enc}, g_{dec}$ on $\mathcal{D}_{his}$, obtaining $\Psi_{\texttt{IsUnseen}}$. (see \refEq{eq.fnd})
    \STATE \textit{/*Pre-train $q_\theta$ using $\mathcal{D}_{his}$*/}
    \REPEAT
        \STATE Sample a batch of episodes $\{\tau_k\}_k$ from $\mathcal{D}_{his}$.
        \STATE Sample a subset of prototypes $I\subseteq P_{seen}$ and an injection $\psi: I\rightarrow P_{unseen}$.
        \STATE Update $q_\theta$ using $\{\tau_k\}_k, f_{I, \psi}$ according to \refEq{eq.inference}.
    \UNTIL{convergence}
    \STATE \textit{/* Train $\pi_{exp}$ using $\mathcal{M}_S$ and $q_\theta$ */}
    \REPEAT
        \STATE Sample $I\subseteq P_{seen}$ and $\psi: I\rightarrow P_{unseen}$.
        \STATE Running the latest $\pi_{exp}$ on $\mathcal{M}_S$ (with $f_{I, \psi}$) to obtain trajectories $\{\tau_k\}_k$
        \STATE Relabel the rewards of $\{\tau_k\}_k$ using  the intrinsic rewards generated by $q_\theta$. (see \refEq{eq.exp_rew})
        \STATE Update $\pi_{exp}$ with PPO using $\{\tau_k\}_k$.
    \UNTIL{certain steps}
\end{algorithmic}
\end{algorithm}

\section{Method}\label{sec.method}
As stated in \refsec{sec.prel}
, we assume the observation space can be divided into the direct product of multiple object representation spaces: $S=\prod_{i=1}^N O$. 
We further assume that each object $o$ has been assigned a category label $o^c$ according to its appearance, which can be obtained by oracle or by unsupervised clustering on objects.

The goal of \alg{} is to learn a prototype mapping function $f_{proto}: O\rightarrow P_{seen}=\{1,2,...,C\}$ that assigns a prototype $o^p$ to each object $o$ in \dsrc{} and \dtarg{} such that objects within the same prototype share the same functionalities. Intuitively, the prototype of an object can represent its functionality, therefore objects with the same prototype can be treated equally no matter which domain (\dsrc{} or \dtarg{}) they come from.

In \dsrc{}, we simply define $f_{proto}$ as $f_{proto}|_S(o)=o^c$ (i.e. $o^p=o^c$) which means prototypes are exactly the category labels of objects. 
This is because objects with the same appearances share the same functionalities.
However, it is not the case in \dtarg{} because we have to map objects into the prototype space \emph{aligned} with \dsrc{} such that our task policy $\pi_{task}$ is applicable. 
An object in \dtarg{} can be seen or unseen depending on whether it has shown in \dsrc{}. 
For the seen object, we can safely apply $f_{proto}|_{S}$ to obtain its prototype. 
For the unseen object, $f_{proto}|_{S}$ is not applicable, 
therefore we have to explore its functionality to determine its prototype. 

\begin{figure}
    \centering
    \includegraphics[width=0.5\textwidth]{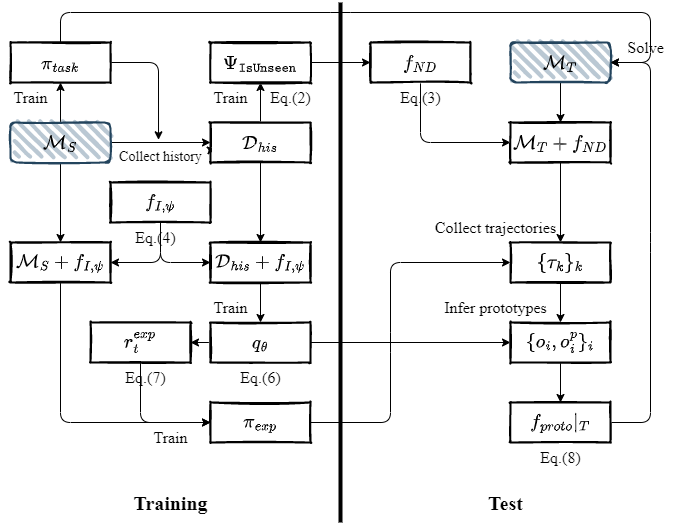}
    \caption{The training and test procedures of \alg{}.}
    \label{fig.arch}
\end{figure}

The overall procedures of \alg{} are presented in Algorithm \ref{alg.opa} and \reffig{fig.arch}.
In the training phase, we first train an indicator $\Psi_{\texttt{IsUnseen}}$ to distinguish unseen objects (\refsec{sec.method.novelty}). Then, we train an exploration policy $\pi_{exp}$ and an inference model 
$q_\theta$ in \dsrc{} (\refsec{sec.method.pa}), which aim to efficiently discover the prototypes of unseen objects. 
In the test phase, we obtain $f_{proto}|_T$ for \dtarg{} by combining $\Psi_{\texttt{IsUnseen}}, \pi_{exp}$ and $q_\theta$ together (\refsec{sec.method.reuse}), with which $\pi_{task}$ can be transferred to \dtarg{}.

\subsection{Novelty Detection}\label{sec.method.novelty}
For an object in \dtarg{}, we want to classify whether it has shown in \dsrc{}. 
This problem is actually a task of novel detection, and many approaches in this field are able to solve it. 
For simplicity, in this work, we consider a native approach that relies on reconstruction loss.

We collect some object samples $O_S=\{o_j\}_{j=1}^M$ from \dsrc{}, and train an auto-encoder (consisting of $g_{enc}$ and $g_{dec}$) that tries to map $o_j\in O_S$ into a latent space via $g_{enc}$, and then map the resulting latent back into $o_j$ via $g_{dec}$. Since the $g_{enc},g_{dec}$ will be over-fitted to the $O_S$, it will present high reconstruction loss if faced with out-of-distribution samples, and therefore can be a hint for unseen objects:
\begin{equation}
    \begin{aligned}
        \Psi_{\texttt{IsUnseen}}(o) = \| g_{dec}\circ g_{enc}(o) - o \|_2 \geq \eta.
    \end{aligned}
    \label{eq.unseen}
\end{equation}
For a seen object in \dtarg{}, we can adopt $f_{proto}|_S$ to obtain its prototype. For an unseen object, we want to remind the agent to explore its functionalities, therefore we also map it into a special prototype space $P_{unseen}$ via an injection $\phi$ ($P_{seen} \cap P_{unseen}=\emptyset$). Therefore, the overall mapping function of novelty detection is:
\begin{equation}
    \begin{aligned}
        f_{ND}(o) = \begin{cases} f_{proto}|_S(o), &\texttt{if not}\ \Psi_{\texttt{IsUnseen}}(o) \\ \phi(o^c), & \texttt{if}\ \Psi_{\texttt{IsUnseen}}(o) \end{cases},
    \end{aligned}
    \label{eq.fnd}
\end{equation}
where $o^c$ is the category label of $o$, and $\phi$ is an injection that maps $o^c$ to $P_{unseen}=\{C+1, ...,2C\}$.
Note that the exact value of $\phi(o^c)$ does not matter because the prototypes of objects in $P_{unseen}$ are all unknown and require to be explored.



\subsection{Online Prototype Alignment}\label{sec.method.pa}
In this section, we aim to train an exploration policy $\pi_{exp}$ that can interact with unseen objects of \dtarg{} (i.e., $\{o:f_{ND}(o)\in P_{unseen}\}$) in a purposeful manner to discover their prototypes.  
However, we have no access to \dtarg{} in the training phase; Even though we do have it, we do not know the real prototypes of unseen objects which are needed for training $\pi_{exp}$. 

Fortunately, we can create some `imaginary' environments from \dsrc{} to train $\pi_{exp}$ in which we have access to ground-truth prototypes via $f_{proto}|_S$. 
At the beginning of an episode, we randomly sample a subset of prototypes $I \subseteq P_{seen}$ and then map them into $P_{unseen}$:
\begin{equation}
    \begin{aligned}
        f_{I, \psi}(o) = \begin{cases} o^p, &\texttt{if}\  o^p\notin I \\ \psi(o^p), &\texttt{if}\  o^p\in I \end{cases},
    \end{aligned}
    \label{eq.fi}
\end{equation}
where $o^p=f_{proto}|_S(o)$ is the prototype of $o$ and $\psi: I\rightarrow P_{unseen}$ is a randomly sampled injection. 
Note that the randomness of $\psi$ is essential, otherwise we can easily infer the prototypes by leveraging  $\psi$, which is actually a backdoor of non-sense.
Both $I$ and $\psi$ keep fixed in the remaining part of the episode. 
Without loss of generality, we further assume the codomain of $\psi$ is $P_{unseen}^I = \{C+1, C+2,..., C+|I|\}$.

Compared \refEq{eq.fi} and \refEq{eq.fnd}, we can see that they induce the same prototype encodings (if we ignore the differences in $P_{unseen}$) when $I=\{o^c:\Psi_{\texttt{IsUnseen}}(o)=\texttt{True}\}$, which means that we can learn $\pi_{exp}$ in \dsrc{} with $f_{I,\psi}$ and then apply it to \dtarg{} with $f_{ND}$. 

The exploration policy $\pi_{exp}$ is trained in \dsrc{} equipped with $f_{I,\psi}$. The aim of $\pi_{exp}$ is to interact with objects in the $P_{unseen}^I$, and $\pi_{exp}$'s behaviour should be informative to infer the original prototypes. To this end, we propose to maximize the mutual information of the trajectory induced by $\pi_{exp}$ (which is denoted as $\tau_{exp}$) and the original prototypes of $P_{unseen}^I$ (which are $I'=[\psi^{-1}(C+1), ...,\psi^{-1}(C+|I|)]$). Formally, $\pi_{exp}$ is trained to maximize the following objective:
\begin{equation}
    \begin{aligned}
        MI(\tau_{{exp}}; I') &= H(I') - H(I'|\tau_{{exp}}) \\
        &\geq H(I') + \mathbb{E}_{\tau \sim I,\psi,\pi_{exp}} \log \ q_\theta(I'|\tau) \\
        &= \mathbb{E}_{\tau \sim I,\psi,\pi_{exp}}\sum_{t=0}^T \log \frac{q_\theta(I'|\tau_{:t+1})}{q_\theta(I'|\tau_{:t})} + Const,
    \end{aligned}
    \label{eq.mi}
\end{equation}
where $q_\theta$ is an inference model that can predict $I'$ given a trajectory,  $\tau_{:t} = [s_0,a_0,r_0,...,s_{t}]$\footnote{$\tau_{:0}:=\emptyset$} is the sub-trajectory consisting of first $t$ transitions in $\tau$. The second line in \refEq{eq.mi} comes from the lower bound proposed in \cite{im_lowerbound}, and the third line follows from the expansion along the time-step dimension and ignores the terms that are not related to $\pi_{task}$. 
Note that $MI(\tau^{{exp}}; I')$ can be maximized by maximizing the lower bound in \refEq{eq.mi}.

To predict $I'$ as soon as possible in an episode, $q_\theta$ is trained using all sub-trajectories $\tau_{:t}$, and the loss function is given as:
\begin{equation}
    \begin{aligned}
        L(\theta) = -\mathbb{E}_{\tau_{:t}\sim I',\psi, \pi_{exp}} \log q_\theta(I'|\tau_{:t}).
    \end{aligned}
    \label{eq.inference}
\end{equation}

To optimize $\pi_{exp}$, we notice that the last line in \refEq{eq.mi} is quite similar to the objective of RL (see \refEq{eq.rl_obj}). Therefore we can maximize \refEq{eq.mi} by giving $\pi_{exp}$ an intrinsic reward as shown in \refEq{eq.exp_rew} and training it using any RL algorithm such as PPO \cite{ppo}:
\begin{equation}
    \begin{aligned}
        r_t^{exp} = \log \frac{q_\theta(I'|\tau_{:t+1})}{q_\theta(I'|\tau_{:t})}.
    \end{aligned}
    \label{eq.exp_rew}
\end{equation}
Intuitively, \refEq{eq.exp_rew} will assign a positive reward to $\pi_{exp}$ if the environment transition at step $t$ (i.e. $(s_t, a_t, r_t, s_{t+1})$) is useful to predict $I'$, which will motivate $\pi_{exp}$ to learn efficient exploration behaviours. These behaviours can reveal the underlying functionalities of unseen elements quickly, therefore are essential for few-shot transfer.

In practice, the modelling of $q_\theta$ and $\pi_{exp}$ is also important because a proper design can introduce useful inductive biases and facilitate the training of $q_\theta$ and $\pi_{exp}$. 
Please refer to Appendix for more details.






\subsection{Policy Reuse}\label{sec.method.reuse}
Our task policy $\pi_{task}$ is built on the prototype space. 
Therefore, we wish to derive $f_{proto}|_T$ that can infer the prototypes in \dtarg{} such that our task policy is applicable when equipped with $f_{proto}|_T$ (i.e. $\pi_{task}\circ f_{proto}|_T$). 

In $\mathcal{M}_T$, we first run $\pi_{exp}$ (with $f_{ND}$ to label unseen elements) for several episodes.
For each episode, we utilize $q_\theta$ to infer the probability distribution of prototypes. We average these distributions to combine them together and then obtain the final prototypes $\{o^p_i\}_i$ of objects $\{o_i\}_i$ based on the aggregate distribution.  
Given $\{(o_i, o_i^p)\}_i$, we train a classifier $f_{cls}$ that can maps $o_i$ to $o_i^p$. In practice, we use PCA and LinearSVC implemented in \cite{sklearn} to realize this classifier because they are light-weighted and run fast. Together with the notations in \refEq{eq.unseen}, our $f_{proto}|_T$ can be formulated as:
\begin{equation}
    \begin{aligned}
        f_{proto}|_T(o) = \begin{cases} f_{proto}|_S(o), &\texttt{if not}\ \Psi_{\Psi_{\texttt{IsUnseen}}}(o) \\ f_{cls}(o), & \texttt{if}\ \Psi_{\texttt{IsUnseen}}(o) \end{cases}.
    \end{aligned}
    \label{eq.f_proto_T}
\end{equation}



\section{Experiment}\label{sec.exp}

\subsection{Environment Setup}
In this work, we mainly consider the task suite Hunter \cite{ocarl} (and also provide results on Crafter \cite{crafter} in the Appendix). 
Hunter is an environment that is designed to be object-centric, which is suitable for our method. It contains 5 kinds of objects in total:  \itembg{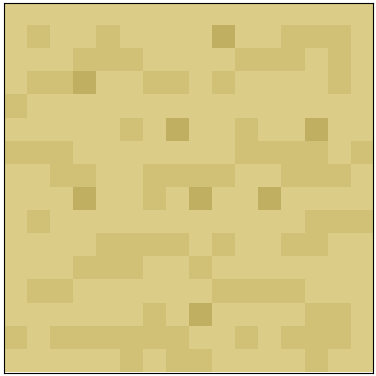}, \itemzombie{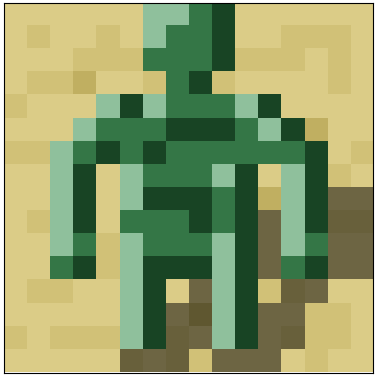}, \itemagent{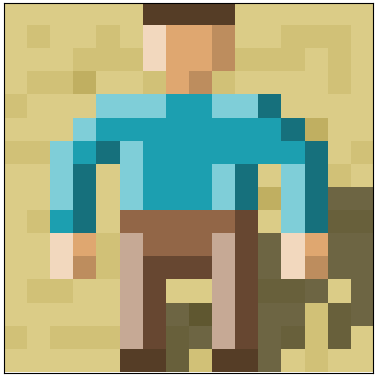}, \itemcow{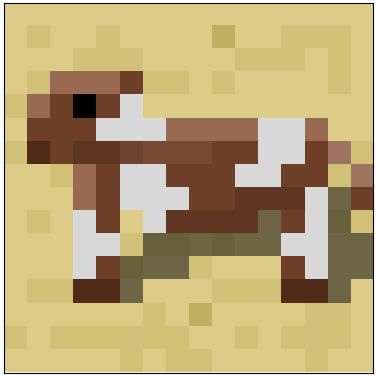} and \itemwall{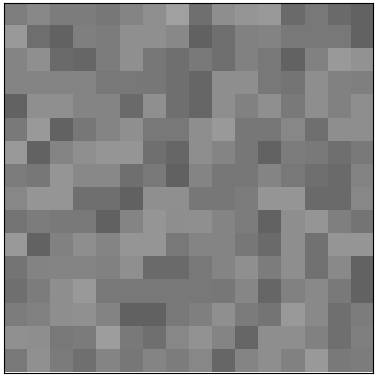}, as shown in \reffig{fig.obs_img} (c). The goal is to train an agent that controls \itemagent{0} to interact with \itemzombie{0} and \itemcow{0}. The same action may result in different rewards when interacting with different objects, e.g., the agent will get a positive reward (=1) if \itemagent{0} shoots at \itemzombie{0}, but a negative reward (=-1) if at \itemcow{0}. Hunter also provides different variants (e.g., Hunter-Z1C1, Hunter-Z2C2,...), which differ in the number of objects.

To test the transfer ability of \alg{}, we derive a new environment from Hunter by changing the appearances of objects (\itembg{0}, \itemzombie{0}, \itemagent{0}, \itemcow{0}, \itemwall{0} $\rightarrow$ \itembg{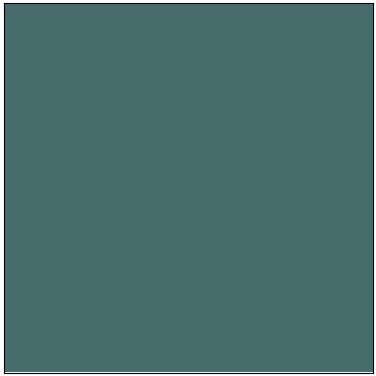}, \itemzombie{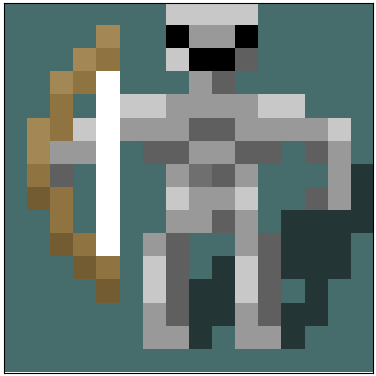}, \itemagent{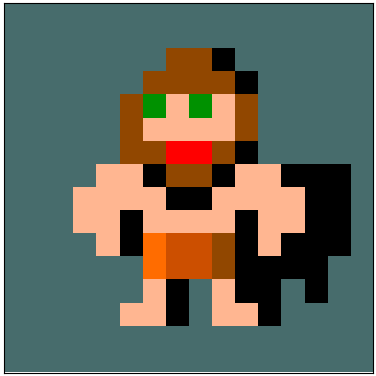}, \itemcow{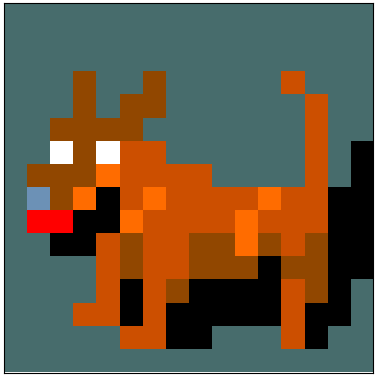}, \itemwall{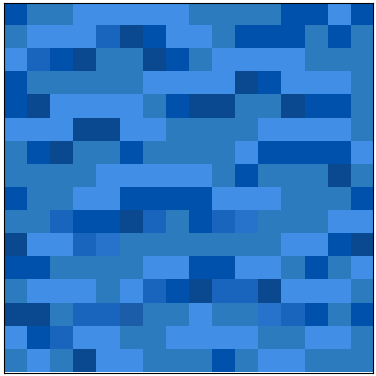}\footnote{These textures come from https://nethackwiki.com/}), as shown in \reffig{fig.obs_img} (d). 
The original and the new environments serve as the source domain and target domain, respectively. To obtain object representations, we divide the $64\times 64\times 3$ image (the observation space in Hunter) into $8\times 8$ tiles, and each tile is of shape $8\times 8 \times 3$. 
By the design of Hunter, each tile contains exactly one object. Therefore, these 64 tiles can be used as object representations for \alg{}. 




\begin{table*}[t]
    \centering
        \caption{The mean and standard deviation of episode returns across 4 seeds, both in the source and target domain. The UNIT4RL(LTMBR)@nM (n=0,3,5) means the UNIT4RL(LTMBR) fine-tuned for n million environment steps in the target domain.}
    \label{tab.all_res}
    \begin{tabular}{c  c c  c c  c c  c c}
    \toprule[1pt]
          & \multicolumn{2}{c}{Hunter-Z1C1} & \multicolumn{2}{c}{Hunter-Z2C2} & \multicolumn{2}{c}{Hunter-Z3C3} & \multicolumn{2}{c}{Hunter-Z4C4} \\
         & Source & Target & Source & Target & Source & Target & Source & Target \\
                  \midrule

        PPO & \ms{1.73}{0.02} & \ms{-0.01}{0.09}& \ms{3.04}{0.28}&\ms{-0.03}{0.13}& \ms{4.15}{0.62}&\ms{-0.04}{0.14}& \ms{5.12}{0.21}& \ms{-0.03}{0.15}\\
        DARLA & \ms{1.25}{0.07} &\ms{-0.01}{0.14} & \ms{1.76}{0.06} & \ms{-0.02}{0.13} & \ms{1.91}{0.09} &\ms{0.01}{0.17}& \ms{2.23}{0.09} &\ms{0.02}{0.2}\\
        LUSR  & \ms{1.14}{0.02} &\ms{-0.33}{0.23} & \ms{1.19}{0.06} & \ms{-0.03}{0.15} & \ms{0.89}{0.23} & \ms{-0.05}{0.19} & \ms{0.90}{0.16}& \ms{-0.03}{0.20}\\
        UNIT4RL@0M &\ms{1.73}{0.02}& \ms{0.22}{1.10} &\ms{3.04}{0.28}&\ms{0.81}{2.12} & \ms{4.15}{0.62} &\ms{-0.87}{0.40}& \ms{5.12}{0.21}& \ms{1.34}{3.31}\\
        LTMBR@0M &\ms{1.50}{0.02}& \ms{0.00}{0.02} &\ms{2.73}{0.03}&\ms{-0.01}{0.04} & \ms{3.89}{0.08} &\ms{-0.01}{0.04}& \ms{4.68}{0.06}& \ms{-0.03}{0.04}\\
        \alg{}(ours) & \ms{1.65}{0.05} & \textbf{\ms{1.71}{0.05}} & \ms{3.22}{0.07}  & \textbf{\ms{3.03}{0.31}} & \ms{4.40}{0.11}  &  \textbf{\ms{4.47}{0.18}} & \ms{5.61}{0.06} & \textbf{\ms{5.68}{0.30}} \\
        \midrule
        UNIT4RL@3M & - & \ms{1.35}{0.33} & - & \ms{3.13}{0.26} & - & \ms{3.84}{0.05}& -&\ms{4.67}{0.77} \\
        UNIT4RL@5M & - & \ms{1.68}{0.05} & - & \ms{3.24}{0.14} & - & \ms{4.35}{0.03}& -&\ms{5.40}{0.4} \\

        LTMBR@3M & - & \ms{1.27}{0.05} & - & \ms{2.14}{0.07} & - & \ms{2.83}{0.07}& -&\ms{3.30}{0.12} \\
        LTMBR@5M & - & \ms{1.38}{0.04} & - & \ms{2.51}{0.10} & - & \ms{3.64}{0.11}& -&\ms{4.63}{0.16} \\
        \bottomrule[1pt]
    \end{tabular}

\end{table*}

\begin{table*}[t]
    \centering
        \caption{The performance ratio of the target and source domain (higher is better) averaged across all environments. Both UNIT4RL and LTMBR need more than 3M adaptation steps in the target domain to match up with \alg{}. }
    \label{tab.perf_ratio}
    \begin{tabular}{c c c c c c | c c}
    \toprule[1pt]
    PPO & DARLA & LUSR &UNIT4RL@0M &LTMBR@0M & \alg{}(ours)  & UNIT4RL@3M  & LTMBR@3M  \\
    \hline
    -0.01 & -0.00 & -0.1 & 0.11 & 0.00 & \textbf{1.00} & 0.91 & 0.84  \\
    \bottomrule
    \end{tabular}

\end{table*}

\subsubsection{Baseline settings}
We compare \alg{} with other approaches designed for domain adaptation, including DARLA\cite{darla}, LUSR\cite{lusr}, UNIT4RL \cite{unit4rl}  and LTMBR \cite{ltmbr}.
DARLA relies on learning disentangled representations to achieve transfer. It utilizes a special $\beta$-VAE in which the reconstruction loss is replaced with a perceptual similarity loss.
LUSR explicitly splits the latent into domain-specific and domain-general features and only relies on domain-general features to build task policy.
UNIT4RL utilizes an image-to-image translation approach named UNIT \cite{unit} that can translate images between domains with unpaired samples. 
When deployed in the target domain, UNIT4RL translates the observations back into the source domain and further fine-tunes the task policy using the translated observations. 
LTMBR introduces an auxiliary task to help the learning of representations in the target domain, which also includes a fine-tuning stage.

All approaches are trained with PPO using the same hyper-parameters. 
For \alg{}, UNIT4RL and LTMBR, we train the task policy $\pi_{task}$ for 25M steps in the source domain. \alg{} uses additional 10M steps in the source domain to train $\pi_{exp}$, and four episodes in the target domain to infer prototypes. 
Since the source domain and target domain are totally different, we also set $I=P_{seen}$ to facilitate the training of $\pi_{exp}$.
For LUSR and DARLA, we find $\pi_{task}$ improves much more slowly, therefore we train $\pi_{task}$ for 100M steps. 
Since UNIT4RL needs observations from the target domain, we collect 0.5M steps in the target domain via a random policy. This dataset is also granted to LUSR \footnote{According to the original paper of LUSR, the data from the target domain is not essential in LUSR if we have access to other variants of environments that are different from the source domain.}. 
For other details, please refer to our Appendix.   

\begin{figure}[t]
    \centering
    \includegraphics[width=0.45\textwidth]{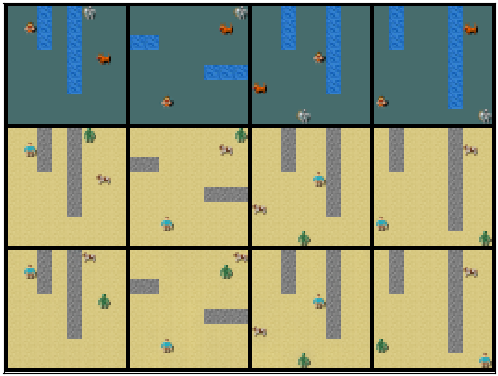}
    \caption{The observations (first row) in the target domain, (second row) generated from the first row using a ground truth mapping function, and (third row) generated using UNIT4RL trained with 4 different seeds.}
    \label{fig.unit_trans}
\end{figure}


\subsubsection{Results}
In \reftab{tab.all_res}, we present the performance results for all baselines. 
Because we are interested in the transfer performance, therefore we also calculate the ratio of performance between the target and source domain ($\texttt{ratio} = \frac{\texttt{performance}(\mathcal{M}_T)}{\texttt{performance}(\mathcal{M}_S)}$) in \reftab{tab.perf_ratio}.
From the results reported in \reftab{tab.all_res} and \ref{tab.perf_ratio}, we can conclude that \alg{} achieves best performance in all tasks.






\begin{figure*}[t]
    \centering
    \includegraphics[width=\textwidth]{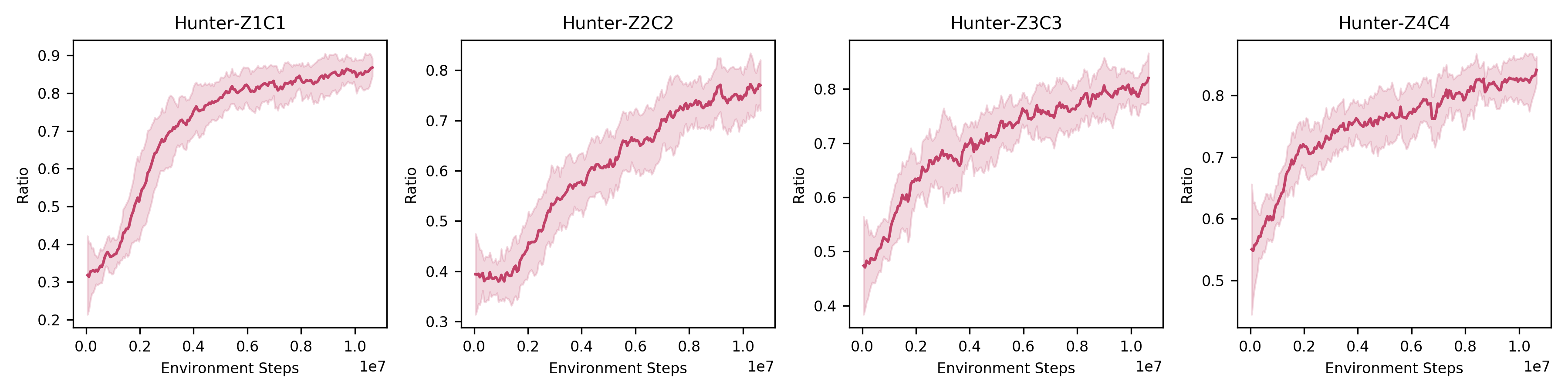}
    \caption{The ratio of episodes that \alg{} can successfully find the ground truth prototype alignment along the training procedure of $\pi_{exp}$. After training, \alg{} can find the ground truth prototypes in a single episode with a probability of more than 0.8.}
    \label{fig.ratio} 
\end{figure*}

\begin{figure*}[t]
    \centering
    \includegraphics[width=\textwidth]{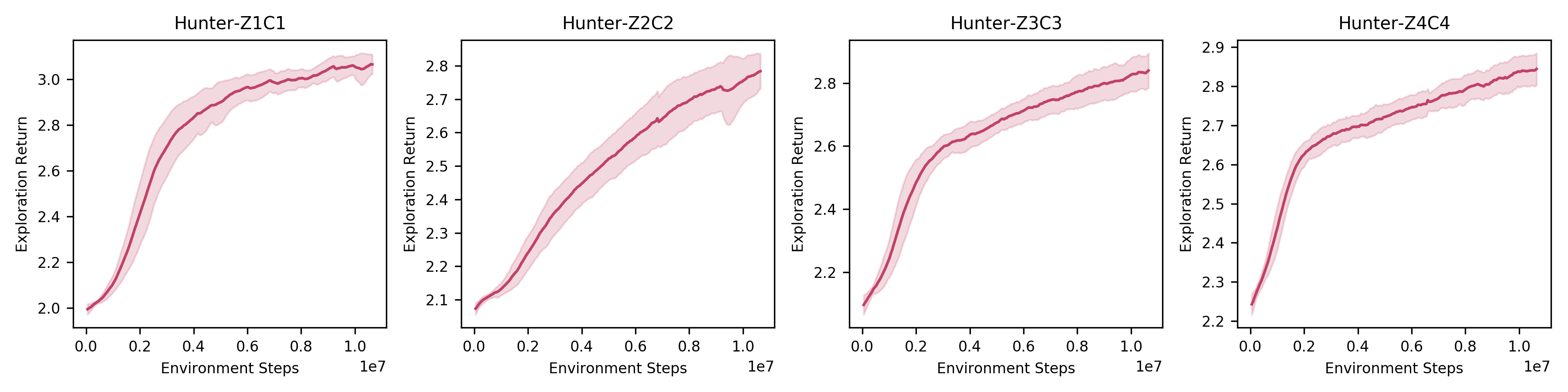}
    \caption{The exploration return produced by the inference model along the training procedure of $\pi_{exp}$. There is an obvious positive correlation between this return and the ratio reported in \reffig{fig.ratio}. }
    \label{fig.expl_ret}
\end{figure*}

In DARLA and LUSR, we find that the $\pi_{task}$ improves much more slowly than other baselines, therefore we train $\pi_{task}$ for 100M steps in the source domain, as we described in the baseline settings. 
However, even with 4x steps, we can still find $\pi_{task}$ can not match up with others. 
We argue that this is because both DARLA and LUSR pre-train an encoder to extract a vectorized latent from observations (and keep frozen in the training of $\pi_{task}$), which ignores the fact that the environments are object-oriented and therefore results in poor performance.

Despite the inferior task performance of DARLA and LUSR in the source domain, they also totally fail to transfer $\pi_{task}$ to the target domain.
For DARLA, this is not surprising because it does not use any additional data from the target domain and is solely trained in the source domain. 
For LUSR, we find that the domain-general and domain-specific features are not well-regularized (see Appendix), in that the domain-specific features can also contain important features such as the position of objects. 
Therefore, the domain-general features may lose important information, which can also explain its inferior task performance compared with DARLA in the source domain.

For UNIT4RL and LTMBR, we further fine-tune $\pi_{task}$ for 3M and 5M steps in the target domain. 
As shown in \reftab{tab.all_res}, both UNIT4RL and LTMBR accelerate the fine-tuning process and only spend less than 5M steps to match up the PPO policy trained for 25M steps. 
Compared with these approaches, \alg{} only needs about 100 steps in the target domain to achieve almost optimal transfer performance,  which is significantly less than the need of UNIT4RL and LTMBR (3M$\sim$5M).

To further expose the failure mode of UNIT4RL@0M and other image-to-images approaches for domain adaptation, we present the translation results of UNIT4RL in \reffig{fig.unit_trans}. We can see that UNIT4RL can discover the mapping between \itembg{3}$\rightarrow$ \itembg{0} and \itemwall{3}$\rightarrow$\itemwall{0} because these objects have unique existence distribution compared with others. However, UNIT4RL fails to reliably learn the mappings between $\itemzombie{3}$, $\itemagent{3}$, $\itemcow{3}$ and $\itemzombie{0}$, $\itemagent{0}$, $\itemcow{0}$ in that each different trial can result in a different mapping. A similar phenomenon should also appear in LUSR, although not explicitly. 
As we have argued before, this is because these objects can not be distinguished solely via visual clues, and therefore we have to rely on their functionalities to learn the mapping, which is one of the main motivations of our work.

\begin{table*}[t]
    \centering
        \caption{The adaptation performance of \alg{} with different number of exploration episodes. \alg{} can achieve high performance (0.8) even with only 1 episode.}
    \label{tab.multi_episode}
    \begin{tabular}{ccccc|c}
    \toprule[1pt]
         & Hunter-Z1C1 & Hunter-Z2C2 & Hunter-Z3C3 & Hunter-Z4C4 & Aggregate Performance Ratio\\
    \hline
        \alg{}@1episode  & 1.41     & 2.21  & 3.48  &4.64 & 0.80\\
        \alg{}@2episodes  & 1.67    & 2.92  & 4.23  &5.44 & 0.96\\
        \alg{}@4episodes  & 1.71    & 3.05  & 4.47  &5.68 & 1.00\\
        \alg{}@16episodes  & 1.68   & 3.12  & 4.45  &5.63 & 1.00\\
    \bottomrule[1pt]
    \end{tabular}

\end{table*}
\begin{table*}[t]
    \centering
    \caption{The necessity of $\pi_{exp}$. We equip \alg{} with different exploration policies (i.e. $\pi_{exp}, \pi_{random}, \pi_{task}$), and run \alg{} for a single episode in the target domain of Hunter-Z1C1. $\pi_{exp}$ is much more efficient for exploration than $\pi_{random}$ and $\pi_{task}$.}
    \label{tab.exp_pol}
    \begin{tabular}{cccc}
    \toprule
         & $\pi_{exp}$ & $\pi_{random}$ & $\pi_{task}$ \\
    \midrule
        Ratio of Correct Mapping                    & \textbf{0.86}  &  0.32 & 0.28\\
        Ratio of Adaptation Performance             & \textbf{0.85}  &  0.21 & 0.23\\
        Average Number of Informative Interactions   & \textbf{1.62}  &  0.12 & 0.08\\
        \bottomrule
    \end{tabular}

\end{table*}

\subsection{Ablation Study}
\subsubsection{The quality of prototype alignment}
To better evaluate the quality of prototypes discovered by \alg{}, in \reffig{fig.ratio} we plot the ratio of episodes that \alg{} can successfully match with the ground truth prototypes of unseen objects in the target domain. 
We can see that this ratio continues to increase during the training process of $\pi_{exp}$ and eventually reaches 0.8+ for all environments. This means \alg{} can find the ground truth prototypes in a single episode with a probability of more than 0.8, which can be further improved by multi-episode exploration.

In \reffig{fig.expl_ret}, we plot the exploration return (produced by $q_\theta$) of $\pi_{exp}$. We can notice that there is an obvious positive correlation between this return and the ratio plotted in \reffig{fig.ratio}. This means that the intrinsic reward generated by $q_\theta$ is informative and instructive because when following this reward $\pi_{exp}$ can improve its ability to find the ground truth prototypes.

In our experiment setting, \alg{} takes four episodes in the target domain for exploration. In \reftab{tab.multi_episode}, we report the performance of \alg{} with other numbers of episodes. We can see that \alg{} achieves a performance ratio of 0.8 even only has access to a single episode in the target domain, and two episodes can quickly improve this ratio to 0.96. This means \alg{} can still obtain prototype assignments of relatively high quality in the absence of enough exploration chances. 




\subsubsection{The necessity of $\pi_{exp}$}
In \alg{}, we put effort into training  $\pi_{exp}$, and one may ask whether $\pi_{exp}$ can pay back. To answer this question, we compare $\pi_{exp}$ with other easy-to-obtain exploration policies in Hunter-Z1C1, which includes a random policy $\pi_{random}$ and the task policy $\pi_{task}$. 

The results are shown in \reftab{tab.exp_pol}. 
We can see that $\pi_{exp}$ is much more efficient for exploration than $\pi_{random}$ and $\pi_{task}$. Note that the performance of $\pi_{task}$ is almost the same with $\pi_{random}$, which means that $\pi_{task}$ can not present meaningful behaviours in the target domain to facilitate the inference of $q_\theta$. 

To further investigate the difference between $\pi_{exp}$ and $\pi_{random}\&\pi_{task}$, in \reftab{tab.exp_pol} we also report the average number of informative interactions in an episode, which includes meaningful interactions between objects that are useful for distinguishing the prototypes and therefore informative for the functionalities of objects. 
As shown in \reftab{tab.exp_pol}, we can see that $\pi_{exp}$ will manage to find these informative interactions, whereas $\pi_{random}$ and $\pi_{task}$ do not present such a purposeful behaviour.





\section{Conclusion}
In this paper, we propose a novel framework named \alg{} that aims to transfer a policy to an unfamiliar environment in a few-shot manner. 
The key of \alg{} is to introduce an exploration mechanism that can purposefully interact with the unseen elements in the target domain. 
By doing so, we can build a mapping function between these unseen elements to seen elements according to their functionalities, and then transfer the policy trained in the source domain to the target domain. 
Our experiments show that \alg{} can not only achieve better transfer performance on tasks in which other baselines fail but also consume much fewer samples from the target domain.

\section*{Acknowledgements}
This work is partially supported by the NSF of China(under Grants 62102399, 61925208, 62002338, 62222214, U22A2028, U19B2019), Beijing Academy of Artificial Intelligence (BAAI), CAS Project for Young Scientists in Basic Research(YSBR-029), Youth Innovation Promotion Association CAS and Xplore Prize.


\bibliography{example_paper}
\bibliographystyle{icml2023}

\newpage
\appendix
\onecolumn

\begin{table}[t]
    \centering
    \caption{The data consumption for OPA and other baselines, both
in the source and target domain. 'Enc.' and 'Expl.' are corresponding to 'Encoder' and 'Exploration' respectively.}
    \label{tab.data_comp}
    \begin{tabular}{c ccc ccc}
    \toprule
         & \multicolumn{3}{c}{Source Domain} & \multicolumn{3}{c}{Target Domain} \\
         & $\pi_{task}$ & $\pi_{exp}$ & Enc. & Fine-tuning & Enc. & Expl.\\
         \midrule
         OPA(ours) & 25M & 10M & - & - & - & $\approx$ 100 \\
         DARLA & 100M & - & 0.5M & - & - & - \\
         LUSR & 100M & - & 0.5M & - & 0.5M & - \\
         UNIT4RL & 25M & - & - & 0-5M & 0.5M & - \\
         LTMBR & 25M & - & - & 0-5M & - & - \\ 
    \bottomrule
    \end{tabular}

\end{table}

\section{Implementation for Baselines}\label{app.impl}

\paragraph{Implementation for PPO} 
The task policies for all approaches included in this work are trained via PPO.
Our PPO implementation is based on Tianshou \cite{tianshou} which is purely based on PyTorch. 
We adopt the hyper-parameters which are shown in \reftab{tab.ppo_para}.

\paragraph{Implementation for DARLA}
For DARLA, we first collect 0.5M samples in the source domain via a random policy. Using these samples, we train a $\beta$-VAE with a grid search over $\beta=0.1, 0.5, 1, 2, 5, 10$. We set $\beta=2$ because it achieves the best results
In the original paper of DARLA, the reconstruction loss of $\beta$-VAE is replaced by a perceptual similarity loss produced by a denoising autoencoder (DAE). 
However, we find the reconstruction loss works better in our case, therefore is used in practice. 

After pre-training the encoder, we then train a task policy $\pi_{task}$ for 100M steps in the source domain based on this encoder. The encoder is frozen during the training of $\pi_{task}$, and will encode the pixel observation into a latent of size 128. The task policy is a 3-layer MLP with hidden sizes 64, and outputs the action probability and value function.

\paragraph{Implementation for LUSR}
LUSR needs a set of different domains to train the encoder. 
In our case, we simply collect 0.5M samples from the source domain and 0.5M from the target domain to train LUSR. The coefficient of the reverse loss in LUSR is grid-searched for 0.1, 0.5, 1, 2, 5, and we find 0.5 works best in our case.  
LUSR splits the latent representations into domain-specific features $z_{s}$ and domain-general $z_g$ features. We also search for the dimensions of both features (including $(|z_s|, |z_g|)=(8, 32), (8, 64), (16, 64), (16, 128), (32, 128)$), and choose $(16, 128)$ in practice. 

After pre-training the encoder, we then train a task policy $\pi_{task}$ for 100M steps in the source domain based on the domain-general features provided by LUSR. The encoder is frozen during the training of $\pi_{task}$, and will encode the pixel observation into a latent of size 128. The task policy is a 3-layer MLP with a hidden size of 64 and outputs the action probability and value function.

\paragraph{Implementation for UNIT4RL}
First, we collect 0.5M samples from the source domain and 0.5M from the target. This data is used to train an image-to-image translation model $T$. All hyper-parameters of UNIT4RL are the same as in the original paper.

We train a task policy $\pi_{task}$ for 25M steps in the source domain.
When deploying $\pi_{task}$ in the target domain, we first translate the observations into the source domain via the translation model $T$, then calculate the action probability and value function using $\pi_{task}$. $\pi_{task}$ is further fine-tuned for 0-5M steps in the target domain via PPO, with $T$ kept fixed.

\paragraph{Implementation for LTMBR}
LTMBR \cite{ltmbr} introduces an auxiliary task to help the learning of representations in the target domain. We conduct a grid search over the coefficients (1,2,4,8,16) of the auxiliary loss in the Hunter-Z2C2 and then apply the optimal coefficient (=4) to other tasks. We train the task policy $\pi_{task}$ for 25M steps with this auxiliary loss in the source domain.

\begin{table}[b]
    \centering
    \caption{PPO hyper-parameters.}
    \label{tab.ppo_para}
    \begin{tabular}{c c}
    \hline
       Hyper-parameter  & Value  \\
       \hline
       Discount factor & 0.9 \\
       Lambda for GAE & 0.95 \\
       Epsilon clip (clip range) & 0.2 \\
       Coefficient for value function loss & 0.5\\
       Normalize Advantage & True \\
       Learning rate & 5e-4 \\
       Optimizer & Adam \\
       Max gradient norm & 0.5 \\
       Steps per collect & 4096 \\
       Repeat per collect & 3 \\
       Batch size & 256 \\
       \hline
    \end{tabular}

\end{table}

\newpage

\section{Implementation for \alg{}}
The implementation is available at \url{https://github.com/albertcity/OPA}.
\subsection{The modelling of $\pi_{exp}$ and $q_\theta$}\label{app.opa_impl.model}
In practice, the modelling of $q_\theta$ and $\pi_{exp}$ is also important because a proper design can introduce useful inductive biases and facilitate the training of $q_\theta$ and $\pi_{exp}$. For simplicity, in the following we assume the prototype $o^p$ of an object $o$ has already been mapped into $P\cup P_{unseen}^I$ via \refEq{eq.fi}.

For $\pi_{exp}$, we use the predicted prototypes of $q_\theta$ as encodings of objects if $o^p\in P_{unseen}$ and $o^p$ if $o^p\notin P_{unseen}$. The predicted results of $q_\theta$ can directly inform $\pi_{task}$ which objects are still unfamiliar to $q_\theta$, and therefore $\pi_{task}$ can learn to interact with them.

For $q_\theta$, we maintain a hidden state $h_p\in R^F$ for each prototype $p$ in $P_{unseen}^I$ ($p=1,..,|I|)$, which summarizes the history of interactions related to $p$. $h_p$ is also tasked to predict the ground truth prototype (i.e. $\psi^{-1}(p)$) via a learnable classifier. 
All $h_p$s are initialized to the \emph{same} hidden states at the beginning of an episode. 
At each transition $(s_t, a_t, r_t, s_{t+1})$, $h_p$ is updated by the following steps:

\paragraph{Broadcasting $h_p$.}
For each objects $o$ in $s_t, s_{t+1}$, we embed $o^p$ into $R^F$ if $o^p\not\in P_{unseen}^I$, which serves as $o$'s encoding . For $o^p \in P_{unseen}^I$, we use $h_p$ instead, because it summarizes the history of interactions related to $p$. This will give us new representations of $s_t$ and $s_{t+1}$, which are denoted as $[\hat{o}_t^1,...,\hat{o}_t^N] \in R^{N\times F}$ and $[\hat{o}_{t+1}^1,...,\hat{o}_{t+1}^N]\in R^{N\times F}$

\paragraph{Processing the transition information.}
In our settings, each object $o$ actually represents a tile in the original observation, therefore we can re-arrange the $[\hat{o}_t^1,...,\hat{o}_t^N]\in R^{N\times F}$ into $[\hat{o}_t^{i,j}]_{i=1}^H{}_{j=1}^W \in R^{H\times W\times F}$ ($N=H\times W$). $\hat{o}_t^{i,j}$ is corresponding to the $(i,j)$'th tile of location $((i-1)\times size_h, (j-1)\times size_w)$ ($size_h, size_w$ is the size of each tile).

 In order to process the transition information, we first concatenate $\hat{o}_t^{i,j}, \hat{o}_{t+1}^{i,j}, a_t, r_t$ together. 
 This will give us $[\Tilde{o}^{i,j}_t]_{i,j} \in R^{H\times W\times (2F+A+R)}$, where $A$ is corresponding to the one-hot embedding of $a_t$ and $R$ corresponding to the embedding of $r_t$. 
 Then we process the resulting features using several convolution layers of (kernel size=3, stride=1, padding=1), which will give us $\Tilde{O}\in R^{H\times W\times F}$. 
 $\Tilde{O}$ can be seen as a latent that summarizes the information of transition $(s_t, a_t, r_t, s_{t+1})$. 
 
\paragraph{Updating $h_p$.}
In order to extract relative information related to $h_p$ from $\Tilde{O}$, we adopt an attention mechanism to get a latent $z_p$ from $\Tilde{O}$. The query vector of this attention is $h_p$, and both the key and value vectors are $\{\Tilde{O}_{i,j,:}: \hat{o}^{i,j}_{t} = p \ or\  \hat{o}^{i,j}_{t+1} = p\}$. In other words, $z_p$ only extracts information from the objects related to $p$. 
After obtaining $z_p$, $h_p$ is then updated via GRU \cite{gru} by taking $z_p$ as the current input.

By the design of $\pi_{exp}$ and $q_\theta$, we can see that the choice of $\psi$ does not influence the inference results of $q_\theta$ and $\pi_{exp}$ (i.e. any $\psi$ will give the same results), which means we can choose a fixed $\psi$ to simplify the training process.

\subsection{Other implementation details}
In \alg{}, we first train $\pi_{task}$ for 25M steps in the source domain. The $\pi_{task}$ uses the $f_{proto}|_S$ as the encoder of objects, and also adopts a self-attention mechanism to model the relations between objects, which is a common practice in OORL \cite{ocarl, rrl}. 

During the training of $\pi_{task}$, we save its trajectories as $D_{his}$. $D_{his}$ is then used to train the indicator $\Psi_{\texttt{IsUnseen}}$ and the inference model $q_\theta$. 
Thanks to the special design of $\pi_{exp}, q_\theta$,  $\psi$ can be fixed for simplicity as explained in the \refapp{app.opa_impl.model}. 

After pre-training $q_\theta$, we train an exploration policy $\pi_{exp}$ for 10M steps in the source domain using the intrinsic rewards generated by $q_\theta$. The network architecture in $\pi_{exp}$ is the same as $\pi_{task}$.


\newpage

\section{Analysis of LUSR}
LUSR splits the latent embedding of an observation $o$ into domain-general and domain-specific features, which are denoted as $z_g(o)$ and $z_s(o)$ respectively. 
Intuitively,  $z_g(o)$ should contain crucial information such as the position of each object, and $z_s(o)$ should contain non-important information such as the image style of the observation that is different in different domains.  

\begin{figure}[t]
    \centering
    \includegraphics[width=\textwidth]{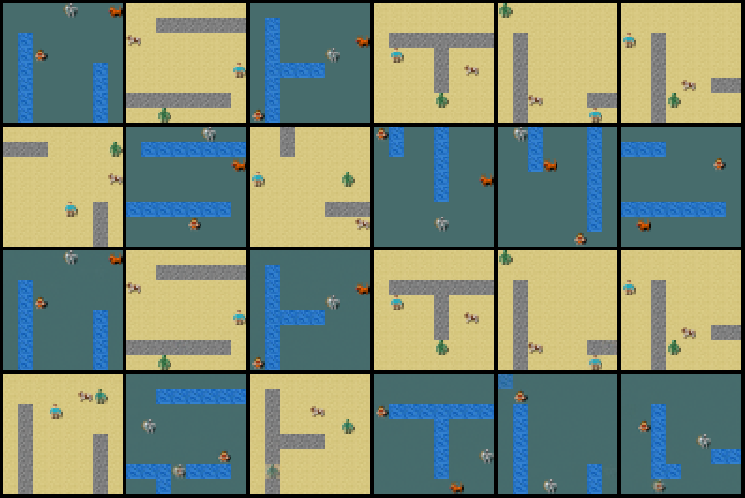}
    \caption{The reconstruction results of LUSR. First and second row: two sampled observations $a,b$; Third row: the reconstruction results of LUSR using $z_g(a)$ and $z_s(a)$; Fourth row: reconstruction using $z_g(a)$ and $z_s(b)$. Although the $z_s(a)$ and $z_s(b)$ are able to distinguish the source and target domain, it also contains important information such as the positions of objects.}
    \label{fig.lusr_enc}
\end{figure}

In \reffig{fig.lusr_enc}, we show the reconstruction results of LUSR using different combinations of $z_g$ and $z_s$. 
We can see that the positions of objects not only depend on $z_g$ but also on $z_s$.
Although the $z_s$ can be used to distinguish the source and target domain, it also contains important information such as the positions of objects. This is problematic because $z_g$  may lose important information.

\newpage
\section{Results on Crafter}
In this section, we provide the transfer results on the Crafter \cite{crafter}, which is a complicated 2-D Minecraft-like environment. We use the original version of Crafter as the source domain. The target domain is a modified version in which we select several objects (i.e. 'stone', 'tree', 'coal', 'cow', 'zombie', 'skeleton') and replace their textures using icons from the Nethack (\url{https://nethackwiki.com/}).

The $\pi_{task}$ are trained for 20M for all algorithms in $\mathcal{M}_S$. For LUSR and UNIT4RL, we take 0.5M from $\mathcal{M}_T$ to train the encoder. For \alg{}, we take 4 episodes to run $\pi_{exp}$ in $\mathcal{M}_T$ to transfer $\pi_{task}$. When training $\pi_{exp}$, we set $I$ to the chosen objects that are different in $\mathcal{M}_S$ and $\mathcal{M}_T$, which can accelerate the training process of $\pi_{exp}$. The observation of Crafter can be rendered into an image of size $72 \times 72 \times 3$, which consists of two parts: (part A) the $7\times 9$ region around the agent (of size $56\times 72 \times 3$ in the image), and (part B) the status of the agent and items in its backpack (of size $16 \ \times 72 \times 3$). When building the encoder for $\pi_{task}, \pi_{exp}, q_\theta$, we separate parts A and B, and use the numerical closed-form of part B (instead of pixels).

The overall results are as shown in \reftab{tab.crafter_res}. According to this table, OPA still achieves the best transfer performance in the Crafter environment. Interestingly, we find that UNIT4RL@20M can not recover the performance in the source domain even after training for 20M in the target domain, which means that the learned mapping function has lost some important information.

\begin{table}[]
    \centering
    \caption{Results on Crafter.}
    \label{tab.crafter_res}
    \begin{tabular}{c|c|c|c}
    \toprule
    Algorithm & Source Domain & Target Domain & Ratio \\
    \midrule
    PPO & \ms{11.62}{0.37}& \ms{3.00}{0.57} & 0.26\\
    DARLA & \ms{7.50}{0.29} & \ms{4.30}{0.37} & 0.57 \\
    LUSR & \ms{7.97}{0.23} &  \ms{2.15}{0.27} & 0.27 \\
    UNIT4RL@0M & \ms{11.62}{0.37} & \ms{3.50}{0.23} & 0.30\\
    \alg{}(ours) & \ms{11.57}{0.52} & \ms{10.69}{0.41} & 1.01 \\
    \midrule
    UNIT4RL@5M & - & \ms{7.88}{0.45} & 0.68 \\
    
    UNIT4RL@20M & - & \ms{9.17}{0.21} & 0.79\\
    \bottomrule    
    \end{tabular}

\end{table}

\newpage
\section{Other Discussions}
\alg{} introduce several stages which may bring cumulative errors. In this section, we analyse this problem.

Formally, there are four parts in \alg{} that may bring errors:
\begin{enumerate}[(1)]
    \item $\Psi_{unseen}: O\rightarrow {0,1}$ in \refEq{eq.unseen},
    \item $q_\theta: \tau\rightarrow I'$ in \refEq{eq.inference},
    \item $\pi_{exp}: S\rightarrow A$,
    \item $f_{cls}: O\rightarrow P_{seen}$ in \refEq{eq.f_proto_T}.
\end{enumerate}

However, the approximation errors brought by (1) and (4) should be very small because the domain of both (1) and (4) is the space of objects $O$, which is simple and small in many cases. In our environment, $O$ is actually a set of size 4. Even in the Crafter (a complicated 2D Mincraft-like environment), it is just 19. Therefore, the error brought by (1) and (4) can be ignored in many cases. For example, $\Psi_{unseen}$ can correctly identify all unseen objects in our cases as we will show later.
The training of (3) relies on (2), therefore the quality of (2) does affect the training of (3). However, this is unavoidable, because (2) and (3) are designed to work together.

\paragraph{The accuracy of $\Psi_{unseen}$.} The binary classifier $\Psi_{unseen}$ is built upon $g_{dec}\circ g_{enc}$(i.e. $\Psi_{unseen}=||g_{dec}\circ g_{dec}(o) - o||\geq \eta$). Therefore, we can use the reconstruction error $||g_{dec}\circ g_{enc}(o) - o||$ to measure the accuracy of $\Psi_{unseen}$. We expect a small error for objects in the source domain and a large error for objects in the target domain. We list the reconstruction error for all objects in \reftab{tab.recon_error}. As shown in \reftab{tab.recon_error}, the reconstruction error is significantly different in the source and target domain, which means that $\Psi_{unseen}$ can correctly identify the unseen objects.

\begin{table}[t]
    \centering
    \caption{The reconstruction errors in the source and target domain.}
    \label{tab.recon_error}
    \begin{tabular}{c c c c c}
    \toprule
     & $obj_1$ & $obj_2$ & $obj_3$ & $obj_4$ \\
    \midrule
    Source Domain (Seen objects) & 0.0112 & 0.0142 & 0.0179 & 0.0410 \\
    Target Domain (Unseen objects) & 4.920 & 8.034 & 9.998 & 14.327\\
    \bottomrule
    \end{tabular}

\end{table}



\end{document}